\documentclass{article}

\PassOptionsToPackage{noend}{algorithmic}

\usepackage[arxiv]{icml2024}

\usepackage[utf8]{inputenc} %
\usepackage[T1]{fontenc}    %
\usepackage{hyperref}       %
\usepackage{url}            %
\usepackage{booktabs}       %
\usepackage{multirow}
\usepackage{amsfonts}       %
\usepackage{nicefrac}       %
\usepackage{microtype}      %
\usepackage{xcolor}         %
\usepackage{blindtext}
\usepackage[colorinlistoftodos,prependcaption]{todonotes}
\usepackage{regexpatch}
\usepackage{orcidlink}
\usepackage{booktabs,colortbl} %
\usepackage{amsmath}
\usepackage{textcomp} %
\usepackage[capitalize,noabbrev]{cleveref}
\usepackage{bm}
\usepackage[inline]{enumitem}
\usepackage{mleftright} %
\usepackage{bbm}  %
\usepackage{subcaption}
\usepackage{lipsum} 
\usepackage{tikz}
\usepackage{xcolor}
\usepackage{placeins} %
\usepackage{csquotes}

\usetikzlibrary{shapes,arrows,positioning}
\definecolor{encodercolor}{RGB}{255, 202, 175}
\definecolor{encoderdarkcolor}{RGB}{218, 184, 148}
\definecolor{losscolor}{HTML}{C6E2E9}
\definecolor{yellowcolor}{RGB}{241, 255, 196}
\definecolor{originaltimeseriescolor}{HTML}{ED85B7}
\definecolor{augtimeseriescolor}{HTML}{218AEA}

\DeclareUnicodeCharacter{00D8}{\O}

\usepackage{amsmath,amsfonts,bm}

\def\eqref#1{equation~\ref{#1}}

\def\1{\bm{1}}

\DeclareMathAlphabet{\mathsfit}{\encodingdefault}{\sfdefault}{m}{sl}
\SetMathAlphabet{\mathsfit}{bold}{\encodingdefault}{\sfdefault}{bx}{n}

\newcommand{\R}{\mathbb{R}}

\newcommand{\denseDots}{\ifmmode\mathinner{\ldotp\kern-0.2em\ldotp\kern-0.2em\ldotp}\else.\kern-0.13em.\kern-0.13em.\fi}
\newcommand{\denserDots}{\ifmmode\mathinner{\kern-0.1em\ldotp\kern-0.2em\ldotp\kern-0.2em\ldotp\kern-0.1em}\else.\kern-0.13em.\kern-0.13em.\fi}

\DeclareMathOperator{\MLP}{MLP}

\DeclareMathOperator{\Beta}{Beta}
\DeclareMathOperator{\Linear}{Linear}
\DeclareMathOperator{\BatchNorm1D}{BatchNorm1D}
\DeclareMathOperator{\ReLU}{ReLU}
\DeclareMathOperator{\similarity}{sim}
\DeclareMathOperator{\CE}{CE}
\newcommand\norm[1]{\lVert#1\rVert}
\newcommand{\TFC}{TF\nobreakdash-C}
\newcommand{\densePar}[1]{\textbf{#1}\hspace{1.5ex}}

\icmltitlerunning{United We Pretrain, Divided We Fail!}

\begin{document}

\twocolumn[
\icmltitle{United We Pretrain, Divided We Fail!\\ Representation Learning for Time Series by Pretraining on 75 Datasets at Once}

\icmlsetsymbol{equal}{*}

\begin{icmlauthorlist}
\icmlauthor{Maurice Kraus}{equal,tuda}
\icmlauthor{Felix Divo}{equal,tuda}
\icmlauthor{David Steinmann}{tuda}
\icmlauthor{Devendra Singh Dhami}{hAI,tue}
\icmlauthor{Kristian Kersting}{tuda,hAI,dfki,ccs}
\end{icmlauthorlist}

\icmlaffiliation{tuda}{Department of Computer Science, TU~Darmstadt, Darmstadt, Germany}
\icmlaffiliation{hAI}{Hessian Center for AI (hessian.AI), Darmstadt, Germany}
\icmlaffiliation{tue}{Mathematics and Computer Science Departement, TU Eindhoven, Eindhoven, Netherlands}
\icmlaffiliation{dfki}{German Research Center for Artificial Intelligence (DFKI), Darmstadt, Germany}
\icmlaffiliation{ccs}{Centre for Cognitive Science, TU~Darmstadt, Darmstadt, Germany}

\icmlcorrespondingauthor{Maurice Kraus}{maurice.kraus@cs.tu-darmstadt.de}
\icmlcorrespondingauthor{Felix Divo}{felix.divo@cs.tu-darmstadt.de}

\icmlkeywords{machine learning, representation learning, pretraining, time series, multi-dataset training}

\vskip 0.3in
]

\printAffiliationsAndNotice{\icmlEqualContribution}

\begin{abstract}
    In natural language processing and vision, pretraining is utilized to learn effective representations.
    Unfortunately, the success of pretraining does not easily carry over to time series due to potential mismatch between sources and target.
    Actually, common belief is that multi-dataset pretraining does not work for time series!
    Au contraire, we introduce a new self-supervised contrastive pretraining approach to learn one encoding from many unlabeled and diverse time series datasets, so that the single learned representation can then be reused in several target domains for, say, classification.
    Specifically, we propose the \emph{XD-MixUp} interpolation method and the \emph{Soft Interpolation Contextual Contrasting} (SICC) loss.
    Empirically, this outperforms both supervised training and other self-supervised pretraining methods when finetuning on low-data regimes.
    This disproves the common belief: We can actually learn from multiple time series datasets, even from 75 at once.
\end{abstract}

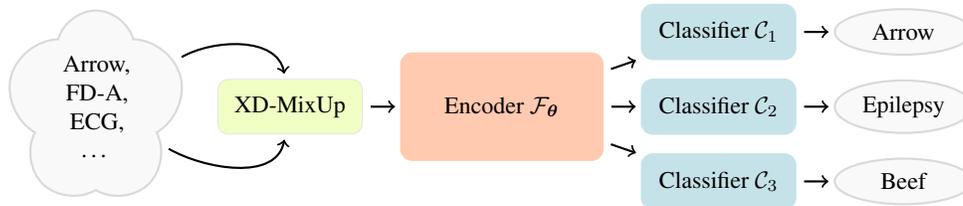
\begin{figure*}[t]
    \begin{center} %
    \begin{tikzpicture}[thick,scale=0.85, every node/.style={scale=0.9}]

   \tikzstyle{classifier} = [
      inner sep=7pt, 
      rectangle, 
      draw, 
      rounded corners, 
      draw=losscolor, 
      fill=losscolor, 
      very thick,
    ]
   \tikzstyle{targetdataset} = [
        ellipse,
        draw,
        text width=1.2cm,
        text centered,
        draw=gray!30,
        fill=gray!5,
        thick,
    ]
    \tikzstyle{encoder} = [
    inner sep=18pt,
    rectangle,
    draw,
    rounded corners,
    draw=encodercolor,
    fill=encodercolor,
    very thick,
    ]
    \tikzstyle{source} = [
        cloud,
        draw,
        text width=1.1cm, %
        align=center,
        cloud ignores aspect,
        thick, 
        draw=gray!30,
        fill=gray!5,
        cloud puffs=5,
        aspect=1
    ]
    \tikzstyle{xdmixup} = [
      inner sep=7pt, 
      rectangle, 
      draw, 
      rounded corners, 
      text centered,
      draw=gray!20, 
      fill=yellowcolor, 
    thin,
    ]

    \tikzstyle{spacing} = [
    shorten >=2pt, shorten <=3pt
    ]

    \node [source] (source) {Arrow,\\FD-A,\\ECG,\\\ldots};
    \node [xdmixup, right=0.5cm of source] (xdmixup) {XD-MixUp};
    \node [encoder, right=0.5cm of xdmixup] (encoder) {Encoder $\mathcal{F}_{\bm\theta}$};
    \node [classifier, right=0.5cm of encoder] (cls2) {Classifier $\mathcal{C}_2$};
    \node [classifier, above=0.25cm of cls2] (cls1) {Classifier $\mathcal{C}_1$};
    \node [classifier, below=0.25cm of cls2] (cls3) {Classifier $\mathcal{C}_3$};
    \node [targetdataset, right=0.5cm of cls2] (target2) {Epilepsy};
    \node [targetdataset, right=0.5cm of cls1] (target1) {Arrow};
    \node [targetdataset, right=0.5cm of cls3] (target3) {Beef};
    
    \draw [->, thick,spacing] (source) to[out=30,in=100]  (xdmixup);
    \draw [->, thick,spacing] (source) to[out=-30,in=-100]  (xdmixup);
    
    \draw [->, thick,spacing] (xdmixup) -- (encoder);
    \draw [->, thick, spacing] (encoder) -- (cls1);
    \draw [->, thick, spacing] (encoder) -- (cls2);
    \draw [->, thick, spacing] (encoder) -- (cls3);
    \draw [->, thick, spacing] (cls1) -- (target1);
    \draw [->, thick, spacing] (cls2) -- (target2);
    \draw [->, thick, spacing] (cls3) -- (target3);
    
    \end{tikzpicture}
    \end{center} %
    \caption{
        The core idea of our method XIT is to learn a single encoder from multiple datasets. The resulting representation can then be used to train classifiers on datasets seen during the pretraining phase and to be transferred to entirely new ones.
    }
    \label{fig:intro}
\end{figure*}

\section{Introduction}
The recent success of large language models~\citep{devlinBERTPretrainingDeep2019,brownLanguageModelsAre2020} as well as diffusion models~\citep{rombachHighResolutionImageSynthesis2022a} has shown that leveraging vast amounts of text and image data can dramatically improve the performance of deep learning models. This has led to impressive advancements in several application areas, such as translation, interactive chat assistants, and text-conditioned image generation. However, there is still a need for a methodology to apply the same principles to the time domain, as significant amounts of \emph{unlabeled} time domain data are available, yet they are seldom used for successfully training models for different tasks. Most classification systems on time series are purely supervised and, therefore, still rely on expensive and complete labels per dataset~\citep{shiSelfSupervisedPretrainingTime2021, nieTimeSeriesWorth2023, zengAreTransformersEffective2023}. Overcoming this is crucial in contemporary real-world situations, such as healthcare, where lack of labels is not the only limitation. There is often an additional overall scarcity of sufficient data points, for instance, due to privacy constraints.
However, this clashes with the requirements of current deep learning models, which typically require large single-source datasets~\citep{iwanaEmpiricalSurveyData2021a}. Fortunately, multiple small datasets exist that, in combination, can be leveraged even if they are unlabeled. For instance, the UCR/UEA Time Series Classification Archive~\citep{dauUCRTimeSeries2019} contains 57\% of datasets with 300 or fewer training samples, which are usually not enough to be applicable to the tasks. Additionally, there are datasets with only unlabeled data, such as the M4 Competition~\citep{makridakisM4CompetitionResults2018} or recordings of meteorological, financial, industrial, traffic, and other signals. Combining them is a new perspective on this collection of valuable data.

We address these challenges by utilizing transfer learning, and specifically, by training a representation on unlabeled source datasets~\citep{eldeleTimeSeriesRepresentationLearning2021}. As shown in \Cref{fig:intro}, the learned features are then used as a starting point for finetuning on a target dataset with typically much fewer labeled instances. In the natural language processing~(NLP) and the vision domain, one can leverage pretrained weights from models trained on a general dataset, even in largely different domains. However, while multiple works have shown the fundamental feasibility of pretraining on time series~\citep{maSurveyTimeSeriesPreTrained2023}, the source and target domain currently still need to be quite similar and follow the same underlying temporal dynamics~\citep{zhangSelfSupervisedContrastivePreTraining2022} for it to succeed. Thus, a similar generalized representation taking advantage of a diverse collection of source datasets would prove very beneficial for use in several target domains.

\pagebreak
\densePar{Contributions}
To this end, we make several important contributions:
\begin{enumerate*}[label=(\textbf{\arabic*})]
    \item We show how up to 75 unlabeled time series datasets can be combined effectively into a single pretraining collection.
    \item We propose a novel interpolation method called Cross-\textbf{D}ataset MixUp~(XD-MixUp) that induces a shared latent representation for multiple datasets.
    \item We propose the \textbf{S}oft \textbf{I}nterpolation \textbf{C}ontextual \textbf{C}ontrasting~(SICC) loss function, which we combine with the Time Series Temporal and Contextual Contrasting~(TS-TCC) \citep{eldeleTimeSeriesRepresentationLearning2021} framework using XD-MixUp. Overall, we call our new architecture XIT\footnote{pronounced <<exit>>}~(\textbf{X}D-MixUp + S\textbf{I}CC + \textbf{T}emporal Contrasting).
    \item We demonstrate good transfer classification performance on multiple small labeled target datasets without requiring extensive retraining for each. In particular, we outperform supervised training and other self-supervised pretraining methods.
\end{enumerate*}

\densePar{Structure of the Paper} We start with explaining our proposed XIT method, including introducing the XD-MixUp and the SICC loss, before moving on to the empirical demonstration of its efficacy. Before concluding, we present related work.

\section{Multi-Dataset Pretraining with XIT}
\begin{figure*}[t]
    \centering
    \includegraphics[width=0.98\textwidth,trim={0.15 0.15cm 0.15 0.15cm},clip]{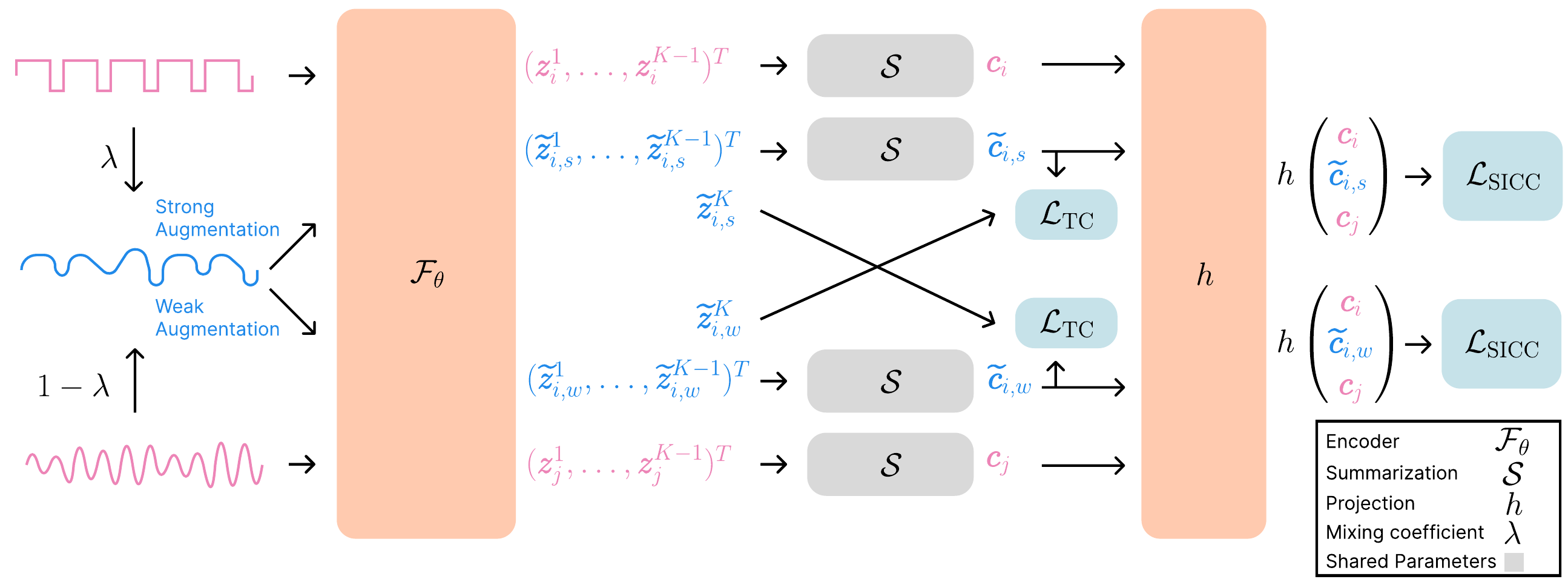}
    \caption{
        Our proposed XIT architecture. From two \textcolor{originaltimeseriescolor}{time series} of a mini-batch, we generate a randomly interpolated variant that gets \textcolor{augtimeseriescolor}{augmented} twice and projected several times along with the original time series. Eventually, we compute two losses to define the overall pretraining objective.
    }
    \label{fig:our_architecture}
\end{figure*}

In this section, we present our pretraining method XIT, where the overall goal is to learn a $D$-dimensional latent representation $\bm{z}_i \in \R^D$ of some time series $\bm{x}_i \in \R^T$ of length $T$. For clarity, we focus on univariate time series, while the method can be readily extended to multivariate tasks. We train the parameters $\bm{\theta}$ of an encoder $\mathcal{F}_{\bm \theta}(\cdot)$ to compress $\bm{x}_i$ into a more abstract representation $\bm{z}_i = \mathcal{F}_{\bm\theta}(\bm{x}_i)$. That representation can subsequently be used for downstream tasks, such as supervised training of a classifier. Note that in the pretraining phase, the encoder $\mathcal{F}$ is trained in a self-supervised fashion without access to any labels. We base our method on the work of~\citet{eldeleTimeSeriesRepresentationLearning2021} and adapt their \emph{TS-TCC} framework to enable the training on multiple datasets. While TS-TCC works well for its designed application to single-dataset scenarios, it cannot leverage the availability of multiple datasets. On the contrary, its performance severely drops in those settings, as we will show in \Cref{sec:experiments:self_supervised}. Similar to their work, we start with a simple 1D convolutional model with three layers as the encoder $\mathcal{F}$. As shown in \Cref{fig:our_architecture}, we merge a pair of time series through interpolation and then derive two augmented variants to calculate two distinct pretraining losses, namely the \textbf{T}emporal \textbf{C}ontrastive~(TC) and \textbf{S}oft \textbf{I}nterpolation \textbf{C}ontextual \textbf{C}ontrastive~(SICC) loss, which we specifically designed to solve the challenges of multi-dataset pretraining and thereby overcome the limitations of TS-TCC. Eventually, the two losses are combined into a single training objective $\mathcal{L}_\text{Total}$ and optimized jointly, yielding the XIT architecture. The complete procedure to perform both pretraining and subsequent finetuning is given in \Cref{alg:combined_procedure}.

\begin{algorithm}[tp]
    \caption{The XIT method consists of the pretraining and finetuning phases.}
    \label{alg:combined_procedure}
    \newcommand{\InlineComment}[1]{\hspace{1ex}\textcolor{gray}{\# #1}}
    \newcommand{\LineComment}[1]{\STATE \textcolor{gray}{\# #1}}
    \begin{algorithmic}
        \STATE \textbf{Input:} Datasets for PT $D_\text{PT}$ (unlabeled) and FT $D_\text{FT}$ (labeled); batch size $B$; constants $\alpha$, $\beta$, and $\tau$; random strong and weak augmentations $A_s$ and $A_w$; initialized models $\mathcal{F}_{\bm\theta}$, $\mathcal{S}$, $g$, and $h$; optimizer $O_\text{PT}$ over all four models; initialized classifier $\mathcal{C}$; optimizer $O_\text{FT}$ over $\mathcal{C}$%
        \STATE \textbf{Output:} learned models $\mathcal{F}_{\bm\theta}$ and $\mathcal{C}$

        \STATE
        \IF{no parameters $\bm\theta$ of $\mathcal{F}$ are known}
            \LineComment{Pretraining phase (loop until convergence)}
            \FOR{mini-batch $\{\bm x_i\}_{i=1}^B \sim D_\text{PT}$}
                \FORALL{$i \in \{1, \ldots, B\}$}
                    \STATE $\bm{z}_i \leftarrow \mathcal{F}_{\bm\theta}(\bm{x}_i)$ \InlineComment{Projections}
                    \STATE $\bm{\kappa}_i \leftarrow h\mleft(\mathcal{S}\mleft(\bm{z}_i^{1:(K-1)}\mright)\mright)$
                    \STATE $\lambda_i \sim \Beta(\alpha, \alpha)$ \InlineComment{XD-MixUp}
                    \STATE $j \leftarrow (i+1) \mod B$
                    \STATE $\bm{\widetilde x}_i \leftarrow \lambda_i \bm x_i + (1 - \lambda_i) \bm x_j$
                    \STATE $\bm{\widetilde x}_{i,s} \sim A_s(\bm{\widetilde x}_i)$ \InlineComment{Strong augmentation}
                    \STATE $\bm{\widetilde z}_{i,s} \leftarrow \mathcal{F}_{\bm\theta}(\bm{\widetilde x}_{i,s})$
                    \STATE $\bm{\widetilde c}_{i,s} \leftarrow \mathcal{S}\mleft(\bm{\widetilde z}_{i,s}^{1:(K-1)}\mright)$
                    \STATE $\bm{\kappa}_{i,s} \leftarrow h(\bm{\widetilde c}_{i,s})$
                    \STATE $\bm{\widetilde x}_{i,w} \sim A_w(\bm{\widetilde x}_i)$ \InlineComment{Weak augmentation}
                    \STATE $\bm{\widetilde z}_{i,w} \leftarrow \mathcal{F}_{\bm\theta}(\bm{\widetilde x}_{i,w})$
                    \STATE $\bm{\widetilde c}_{i,w} \leftarrow \mathcal{S}\mleft(\bm{\widetilde z}_{i,w}^{1:(K-1)}\mright)$
                    \STATE $\bm{\kappa}_{i,w} \leftarrow h(\bm{\widetilde c}_{i,w})$
                \ENDFOR

                \LineComment{Compute the TC loss}
                \STATE $\mathcal{L}_\text{TC}^s = - \frac{1}{B}\sum_{i=1}^B \log \mleft(
                    \frac{g \mleft( \bm{\widetilde c}_{i,w}, \bm{\widetilde z}_{i,s}^K \mright)}{\sum_{j=1}^B g \mleft( \bm{\widetilde c}_{i,w}, \bm{\widetilde z}_{j,s}^K \mright)}
                \mright)$
                \STATE $\mathcal{L}_\text{TC}^w = - \frac{1}{B}\sum_{i=1}^B \log \mleft(
                    \frac{g \mleft( \bm{\widetilde c}_{i,s}, \bm{\widetilde z}_{i,w}^K \mright)}{\sum_{j=1}^B g \mleft( \bm{\widetilde c}_{i,s}, \bm{\widetilde z}_{j,w}^K \mright)}
                \mright)$
                \STATE $\mathcal{L}_\text{TC} \leftarrow \frac{1}{2} \left( \mathcal{L}_\text{TC}^s + \mathcal{L}_\text{TC}^w \right )$

                \LineComment{Compute the SICC loss}
                \STATE $\mathfrak{B}^s = \left( \bm{\kappa}_{1,l}, \denserDots, \bm{\kappa}_{B,l}, \bm{\kappa}_{1,s}, \denserDots, \bm{\kappa}_{B,s}, \bm{\kappa}_{1,r}, \denserDots, \bm{\kappa}_{B,r} \right)$
                \STATE $\mathfrak{B}^w = \left( \bm{\kappa}_{1,l}, \denserDots, \bm{\kappa}_{B,l}, \bm{\kappa}_{1,w}, \denserDots, \bm{\kappa}_{B,w}, \bm{\kappa}_{1,r}, \denserDots, \bm{\kappa}_{B,r} \right)$
                \STATE Compute $\mathcal{L}_\text{SICC} \mleft( \mathfrak{B}^s \mright)$ and $\mathcal{L}_\text{SICC} \mleft( \mathfrak{B}^w \mright)$ by Eq.~\ref{eq:l_sicc_B}
                \STATE $\mathcal{L}_\text{SICC} \leftarrow \frac{1}{2} \left(\mathcal{L}_\text{SICC} \mleft( \mathfrak{B}^s \mright) + \mathcal{L}_\text{SICC} \mleft( \mathfrak{B}^w \mright) \right)$
    
                \STATE $\mathcal{L}_\text{Total} \leftarrow \beta \mathcal{L}_\text{TC} + (1-\beta)\mathcal{L}_\text{SICC}$
                \STATE Update all model parameters with $O_\text{PT}$ to minimize $\mathcal{L}_\text{Total}$
            \ENDFOR
            \STATE Store learned parameters $\bm\theta$ of $\mathcal{F}$
        \ENDIF

        \STATE
        \LineComment{Finetuning phase (loop until convergence)}
        \FOR{mini-batch $\{(\bm x_i, y_i)\}_{i=1}^B \sim D_\text{FT}$}
            \STATE $\bm{z}_i \leftarrow \mathcal{F}_{\bm\theta}(\bm{x}_i)$
            \STATE $\bm{\widehat y}_i \leftarrow \mathcal{C}(\bm{z}_i)$ \InlineComment{Obtain class probabilities}
            \STATE $\mathcal{L}_\text{Class} \leftarrow \CE(\bm{\widehat y}_i, y_i)$ \InlineComment{Use cross-entropy as criterion}
            \STATE Update model parameters with $O_\text{FT}$ to minimize $\mathcal{L}_\text{Class}$
        \ENDFOR
        \STATE \textbf{return} learned composed model $\mathcal{F}_{\bm\theta} \circ \mathcal{C}$
    \end{algorithmic}
\end{algorithm}

\subsection{XD-MixUp and Data Augmentation}
We now construct a pretraining task to train the joint encoder $\mathcal{F}$. Given two different time series $\bm{x}_i$ and $\bm{x}_j$, we first generate a pointwise interpolated variant $\bm{\widetilde x}_i \in \R^T$ by
\begin{equation}
    \label{eq:interpolation}
    \bm{\widetilde x}_i = \lambda_i \bm x_j + (1 - \lambda_i) \bm x_k, \hspace{0.25cm} \lambda_i \sim \Beta(\alpha, \alpha) .
\end{equation}
Here, $\alpha > 0$ is the shape parameter of the symmetric $\Beta$ distribution.
Since $\lambda_i \in [0, 1]$, $\bm{\widetilde x}_i$ is a convex combination of $\bm x_j$ and $\bm x_k$. We will later show how to select them appropriately. This approach is inspired by the label smoothing method MixUp~\citep{zhangSelfSupervisedContrastivePreTraining2022}, which is known to improve robustness to outliers, training stability, and calibration~\citep{thulasidasanMixupTrainingImproved2019} in the image domain and has already been applied to time series by \citet{wickstromMixingContrastiveLearning2022}.

Besides the usual benefits of data augmentation, we use it to generate data points between the different clusters of time series induced by training on multiple datasets, typically each consisting of one or many sub-clusters. This is especially relevant since the eventual downstream task might consist of some time series not seen in the pre-training phase and, therefore, might lie outside the clusters.

A mini-batch consists of randomly sampled time series from each of the included datasets. We select the pairs for interpolation from a single mini-batch of size $B$ on the fly while optimizing. We build distinct pairs by interpolating between consecutive pairs of time series, like $\left( \bm{x}_1, \bm{x}_2 \right), \left( \bm{x}_2, \bm{x}_3 \right), \dots, \left( \bm{x}_B, \bm{x}_1 \right)$ with independently sampled $\lambda_i$. We are then left with a batch of $B$ interpolated time series $\bm{\widetilde x}_i$, where $i \in \{1,\dots,B\}$, in addition to the original time series. We deliberately do not consider \emph{all} possible combinations within a mini-batch to not dramatically increase the batch size, which would incur high computational costs in the loss computations. Following TS-TCC, we then apply a strong and a weak augmentation to our specifically crafted time series $\bm{\widetilde x}_i$ to obtain $\bm{\widetilde x}_{i,s}$ and $\bm{\widetilde x}_{i,w}$.

\subsection{Pretraining Loss}
The strongly and weakly augmented samples are then used to compute the loss $\mathcal{L}_\text{Total}$, which consists of a weighted combination of the temporal contrastive loss $\mathcal{L}_\text{TC}$ and the soft interpolation contextual contrastive loss $\mathcal{L}_\text{SICC}$:
\begin{equation}
    \label{eq:total_loss}
    \mathcal{L}_\text{Total} = \beta \mathcal{L}_\text{TC} + (1-\beta)\mathcal{L}_\text{SICC} .
\end{equation}
The weight $\beta \in [0,1]$ is determined by hyperparameter search. The TC loss guarantees that a time series representation captures its unique, relevant features and differentiates from others. The SICC loss in combination with MixUp ensures meaningful connections between the various datasets seen during pretraining by enforcing a well-structured latent space between them.

As shown in \Cref{fig:our_architecture}, the next step after XD-MixUp is to encode each time series $\bm{\widetilde x}_{i,s}$ into a sequence of $K$ embedding vectors $\left( \bm{z}_{i,s}^1, \dots, \bm{z}_{i,s}^K \right) = \bm{z}_{i,s} = \mathcal{F}(\bm{x}_{i,s}, \bm \theta)$ of size $Z$ each using the convolutional encoder $\mathcal{F}$, where $\bm{z}_{i,s}^k \in \R^Z$. The time series $\bm{\widetilde x}_{i,w}$ and the original $\bm{x}_i$ are encoded into $\bm{\widetilde z}_{i,w}$ and $\bm{z}_i$, respectively. We then use a learned summarization model $\mathcal{S}(\cdot)$ to condense the first $K-1$ embedding vectors $\bm{z}^{1:(K-1)}$ into a single context $\bm{c} \in \R^C$, for all time series individually. Working with these summary contexts instead of with the individual embedding vectors greatly simplifies the computation of the pretraining losses and makes it more efficient. In addition, it promotes learning more high-level features that can compress all embedding vectors into smaller representations~\citep{eldeleTimeSeriesRepresentationLearning2021}. We will now provide an in-depth explanation of both TC and SICC loss.

\subsubsection{Temporal Contrasting}
The TC loss~\citep{eldeleTimeSeriesRepresentationLearning2021} is computed by solving a cross-forecasting task. The context derived from the weakly augmented embedding $\bm{\widetilde c}_w$ is used to predict the last embedding vector $\bm{\widetilde z}_s^K$ of the strong embedding, and vice versa. This is done using a similarity measure $g \mleft( \bm{\widetilde c}, \bm{\widetilde z}^K \mright) = \exp\mleft( \bm{\widetilde c}^T \bm W \bm{\widetilde z}^K \mright)$ with jointly learned $\bm W \in \R^{C \times Z}$. The task is to maximize the similarity to the differently augmented time series embedding of the same time series while minimizing the similarity to the other embeddings from the mini-batch. This favors representations that are invariant to the augmentations being applied. Overall, we compute the loss as follows:
\begin{align*}
    \mathcal{L}_\text{TC}^s &= - \frac{1}{B}\sum_{i=1}^B \log \mleft(
        \frac{g \mleft( \bm{\widetilde c}_{i,w}, \bm{\widetilde z}_{i,s}^K \mright)}{\sum_{j=1}^B g \mleft( \bm{\widetilde c}_{i,w}, \bm{\widetilde z}_{j,s}^K \mright)}
    \mright)
    \\ %
    \mathcal{L}_\text{TC}^w &= - \frac{1}{B}\sum_{i=1}^B \log \mleft(
        \frac{g \mleft( \bm{\widetilde c}_{i,s}, \bm{\widetilde z}_{i,w}^K \mright)}{\sum_{j=1}^B g \mleft( \bm{\widetilde c}_{i,s}, \bm{\widetilde z}_{j,w}^K \mright)}
    \mright)
\end{align*}
We then average them to obtain $\mathcal{L}_\text{TC} = \frac{1}{2} \left( \mathcal{L}_\text{TC}^s + \mathcal{L}_\text{TC}^w \right )$. This loss is called InfoNCE (from \textbf{N}oise \textbf{C}ontrastive \textbf{E}stimation) since \citet{oordRepresentationLearningContrastive2019} have shown that it optimizes a lower bound on the mutual information $I(\bm{\widetilde c} ; \bm{\widetilde z})$ between the context vectors $\bm{\widetilde c}$ and corresponding embeddings $\bm{\widetilde z}$.

\subsubsection{Soft Interpolation Contextual Contrasting}

Our novel SICC loss aligns the information in the augmented time series context vector to the non-augmented contexts. This ensures that the encoder $\mathcal{F}$ remains invariant to the chosen augmentations, allowing it to capture abstract concepts that can be applied to multiple datasets.
We want the context to contain the information of the source time series pair $(\bm x_i, \bm x_j)$ used to form the interpolated time series $\bm{\widetilde x_i}$ depending on the interpolation coefficient $\lambda_i$. Namely, if $\lambda \approx 0$ then $(\bm{\widetilde x_i}, \bm x_i)$ should be a positive pair and $(\bm{\widetilde x_i}, \bm x_j)$ a negative pair. If $\lambda \approx 1$, then the positive/negative relations switch. We, therefore, extend the normal notion of \emph{hard} positive-negative examples to a \emph{soft} variant that treats pairing within the interpolated time series group proportional to $\lambda$ and still considers the rest of the mini-batch to be hard negative samples. Our approach thereby differs from the loss of \citet{sohnImprovedDeepMetric2016} and \citet{chenSimpleFrameworkContrastive2020} used in TS-TCC, where the contextual alignment is solely performed between the two augmented time series. Our approach allows learning representations that are aligned across multiple datasets.

To enforce the entire embeddings $\bm z$ to be aligned, we directly use the contexts $\bm c$ that we computed with the summarization model $\mathcal{S}$. Since we want the information content of a positive pair to match but do not require the concrete representation to be exactly the same, we further project the contexts $\bm c$ using a two-layer learned $\MLP$, obtaining $\bm \kappa = h(\bm c) = \Linear( \ReLU ( \BatchNorm1D ( \Linear (\bm c) ) ) )$. Here, $\Linear(\bm \xi) = \bm W \bm \xi + \bm b$ such that the resulting vector has half the dimension of the input vector, i.e., that $\bm\kappa \in \R^{C / 4}$.
Let $(\bm{\kappa}_{i,l}, \bm{\kappa}_{i,s}, \bm{\kappa}_{i,r})$ with $i \in \{1,\dots,B\}$ be the mini-batch of $B$ triples consisting of either the strongly or weakly augmented time series projection $\bm{\kappa}_i^s$ or $\bm{\kappa}_i^w$ along with the projection of the left $\bm{\kappa}_{i,l}$ and right time series $\bm{\kappa}_{i,r}$ that were interpolated between with $\lambda_i$ to form the augmented one in \cref{eq:interpolation}. We arrange these into two sets of vectors of length $3B$:
\begin{align*}
    \mathfrak{B}^s &= \left( \bm{\kappa}_{1,l},\, \denseDots,\, \bm{\kappa}_{B,l}, \bm{\kappa}_{1,s},\, \denseDots,\, \bm{\kappa}_{B,s}, \bm{\kappa}_{1,r},\, \denseDots,\, \bm{\kappa}_{B,r} \right) ,
    \\
    \mathfrak{B}^w &= \left( \bm{\kappa}_{1,l},\, \denseDots,\, \bm{\kappa}_{B,l}, \bm{\kappa}_{1,w},\, \denseDots,\, \bm{\kappa}_{B,w}, \bm{\kappa}_{1,r},\, \denseDots,\, \bm{\kappa}_{B,r} \right) .
\end{align*}
The loss function is then computed as the average $\mathcal{L}_\text{SICC} = \frac{1}{2} \left(\mathcal{L}_\text{SICC} \mleft( \mathfrak{B}^s \mright) + \mathcal{L}_\text{SICC} \mleft( \mathfrak{B}^w \mright) \right)$, where the individual parts are defined as:
\begin{equation}
    \label{eq:l_sicc_B}
    \begin{aligned}
        \mathcal{L}_\text{SICC} \mleft( \mathfrak{B} \mright)
        =
        \frac{1}{B} \sum_{i=1}^B \Big[ \;
            &\ell( \mathfrak{B}, i, B+i, 1-\lambda_i )
            +
            \\[-11pt]
            &\ell( \mathfrak{B}, B+i, i, 1-\lambda_i )
            +
            \\[-1pt]
            &\ell( \mathfrak{B}, B+i, 2B+i, \lambda_i )
            +
            \\[-5pt]
            &\ell( \mathfrak{B}, 2B+i, B+i, \lambda_i )
        \; \Big],
    \end{aligned}
\end{equation}
\begin{equation}
    \label{eq:ell_sicc}
    \ell \mleft( \mathfrak{B},i,j,\mu \mright) =
    - \log \mleft(
        \frac{
            \exp( \mu \similarity(\mathfrak{B}_i, \mathfrak{B}_j))
        }{
            \sum_{k=1}^{3B} \mathbbm{1}_{k \neq i}
            \exp( \similarity(\mathfrak{B}_i, \mathfrak{B}_k))
        }
    \mright).
\end{equation}
Here, $\similarity(\bm x, \bm{x}') = \frac{\bm{x}^T \bm{x}'}{ \tau \norm{\bm x}_2 \norm{\bm{x}'}_2 }$ denotes the scaled cosine similarity with temperature parameter $\tau$, $\mathbbm{1}$ is an indicator variable that is $1$ if $k \neq i$ and $0$ otherwise. In \Cref{eq:interpolation}, $\lambda_i$ scaled the \emph{distance} of $\bm{\widetilde x}_i$ to $\bm x_i$ and $1-\lambda$ the distance to the other time series, $\bm x_j$. Since we now want to scale \emph{similarities} proportionally, we have to reverse the roles of $\lambda$ and $1-\lambda$. In \cref{eq:ell_sicc}, since we minimize the negative of the fraction, we optimize for maximizing the numerator~(i.e., the similarity of the positive pair $(i,j)$) and minimizing the denominator~(i.e., the similarity of all other negative pairs $(i,k)$). The computation of $\ell$ can be numerically stabilized using the log-sum-exp trick.

We have now defined the three core components of our approach and how they connect. First, we interpolate between datasets with XD-MixUp, to then apply the TC and SICC losses. Together, they form the complete XIT pretraining procedure, whose efficacy we now assess in the next section.

\section{Experiments}
After having laid out the motivation and formal foundation of XIT, we will answer these key research questions with our experiments:
\begin{enumerate*}[label=\textbf{(Q{\arabic*})}]
    \item Do multiple datasets help to learn a more general representation that is easier to transfer to seen and unseen datasets?
    \item Is an encoder pretrained in a self-supervised fashion more helpful than directly learning a classifier, especially for small target datasets?
    \item Which key components of our proposed XIT procedure cause the improvements that we observe?
    \item How discriminative are the learned representations regarding inter- and intra-dataset structures?
\end{enumerate*}
The remainder of this section will provide an overview of our experimental setup, with further details provided in \Cref{sec:app:experimental_setup}. Subsequently, we will present the results and discoveries.

\densePar{Datasets and Augmentations}
We compare our method with the baselines on a  diverse set of classification datasets from the large \emph{UCR Time Series Classification} repository~\citep{dauUCRTimeSeries2019}. At the time of writing, it contained 128 datasets of a wide range of domains, time series lengths, sample counts, and numbers of classes. We used all datasets of up to 600 time steps, resulting in 100 datasets as shown in \Cref{tab:ucr_dataset_distribution} in the Appendix. We did not include longer ones to limit the discrepancy with shorter datasets, which would contain increasingly more padding. In the case of multivariate time series, we exclusively used the first variate for both pretraining and classification. For a level comparison, we ensure that all time series datasets are equal in length by prepending zeros. We use the same augmentations as in the supplementary TS-TCC implementation: magnitude scaling as weak augmentation and \emph{permutation-and-jitter} as strong augmentation.

\densePar{Baselines}
We compare XIT against multiple baselines to evaluate its effectiveness. To this end, we employed the following five self-supervised pretraining methods:
\begin{enumerate*}[label=(\arabic*)]
    \item TS-TCC~\citep{eldeleTimeSeriesRepresentationLearning2021}, which uses temporal and contextual contrasting between weakly and strongly augmented time series to learn an embedding.
    \item \TFC{}~\citep{zhangSelfSupervisedContrastivePreTraining2022}, which learns and aligns a time and frequency domain encoding into a joint representation, again with contrastive methods.
    \item TS2Vec~\citep{yueTS2VecUniversalRepresentation2022}, which performs masking and contrasting hierarchically.
    \item TNC~\citep{tonekaboniUnsupervisedRepresentationLearning2020} aligns neighboring windows with their respective latent representation.
    \item T-Loss~\citep{franceschiUnsupervisedScalableRepresentation2019a} utilizes time-based negative sampling in conjunction with a triplet loss to learn an encoding.
\end{enumerate*}
Additionally, the \emph{supervised} model performs no pretraining, i.e., it uses a randomly initialized encoder that is then used for finetuning.

\densePar{Evaluation} The primary objective of the evaluation is to determine the effectiveness of the representations learned by $\mathcal{F}_{\bm\theta}$.
We therefore follow the \emph{linear probing} experiment of \citet{oordRepresentationLearningContrastive2019}, i.e. learning a shallow neural classifier head $\mathcal{C}$ with a single linear layer while the encoder $\mathcal{F}_{\bm\theta}$ is frozen. The limited capacity of the classifier heads in linear probing allows for a better evaluation of the underlying representations than more powerful classifiers like SVMs or deep neural networks. We evaluate the classification mainly with AUROC and macro-averaged F1 scores since the accuracy metric is unreliable in the face of uneven class distributions (cf. \Cref{tab:ucr_dataset_distribution} in the appendix). %

\subsection{Self-supervised Pretraining on Multiple Datasets:~(Q1) \&~(Q2)}
\label{sec:experiments:self_supervised}

\begin{figure}[t]
    \centering
    \begin{subfigure}[b]{0.95\linewidth}
        \centering
        \includegraphics[width=\linewidth,trim={0 0.25cm 0 0.25cm},clip]{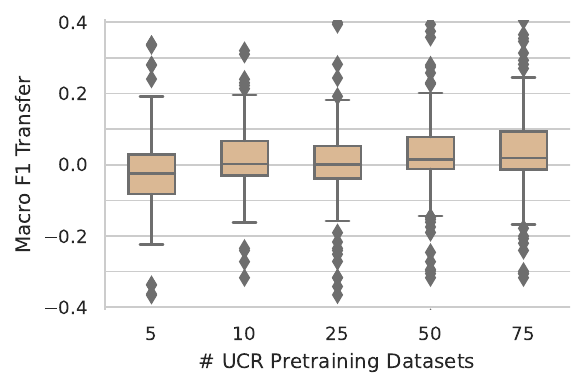}
        \caption{Finetuning on the 100 UCR datasets.}
        \label{fig:exp_transfer_to_many:within}
    \end{subfigure}
    \\
    \vspace{0.5cm}
    \begin{subfigure}[b]{0.95\linewidth}
        \centering
        \includegraphics[width=\linewidth,trim={0 0.25cm 0 0.25cm},clip]{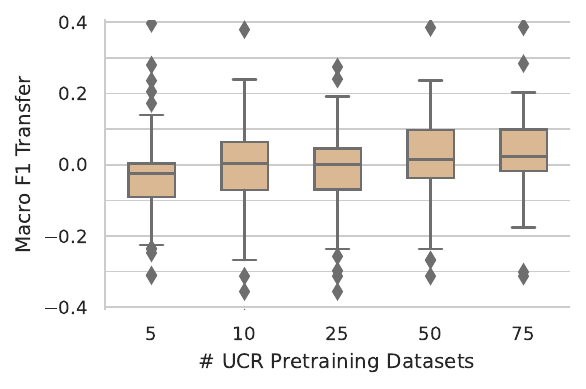}
        \caption{Finetuning on a hold-out set of 25 datasets each.}
        \label{fig:exp_transfer_to_many:hold_out}
    \end{subfigure}
    \caption{
        This box plot shows the transfer surplus over supervised training measured in Macro F1 score difference after pretraining XIT on increasingly large subsets of the UCR repository with three folds each. Higher is better.
    }
    \label{fig:exp_transfer_to_many}
\end{figure}

To determine whether we can learn a single encoder on multiple datasets, we pretrain time series classification models on large portions of the UCR repository. Here, 100 datasets have a fixed sequence length of up to 600. We pretrained on up to 75 different datasets, where we subsequently finetuned both on all 100 datasets~(\Cref{fig:exp_transfer_to_many:within}) and 25 of the held-out ones~(\Cref{fig:exp_transfer_to_many:hold_out}). We sample each domain with the same frequency~(see \Cref{tab:ucr_dataset_distribution} in the Appendix). Within each domain, we sampled time series data points for training according to the size of the respective origin datasets.
We create three folds of each dataset selection, where we each perform the pretraining and finetuning steps with independently initialized models.
We first observe that in \Cref{fig:exp_transfer_to_many:within}, increasing the number of datasets in the pretraining phase increases the benefit of pretraining over supervised training since the Macro F1 transfer becomes increasingly positive.
The effect is even more pronounced when only transferring to entirely unseen datasets, where the increased diversity aids in learning more general representations. In the case of pretraining on 75 datasets, XIT outperforms direct supervised training in 64.7\% of all datasets, while in the hold-out evaluation, this is still true for an impressive 62.5\% of datasets. In conclusion, XIT effectively learns time series representations from multiple, diverse datasets.

To further position XIT among existing approaches, we compare it to five popular baselines. We assess the performance on 25 difficult UCR datasets that were not used for pretraining, in addition to the other 75 datasets.
It is crucial to emphasize, as the other methods have demonstrated limitations in this context, that the conventional pretraining approaches do not work effectively for this scenario. 
\Cref{tab:ucr_comparison_all_baselines} shows the competitive and state-of-the-art performance of XIT. For ease of comparison, we follow \citet{demsarStatisticalComparisonsClassifiers2006} and append ranks for each method. In general, we outperform the purely constrastive methods (TS-TCC, \TFC{}, TNC, and T-Loss). Even when compared to the mixed reconstructive-contrastive method TS2Vec, we outperform it overall.

These experiments show that---in contrast to the findings of~\citet{zhangSelfSupervisedContrastivePreTraining2022}---pretraining on more than one dataset is indeed feasible and beneficial when compared to supervised given an appropriate model is employed.

\begin{table*}[t]
    \centering
    \caption{
        This table shows the classification performance of XIT in comparison to baselines when pretrained on 75 datasets of the UCR repository. The shown Macro F1 scores were computed on three independent seeds and 25 datasets not seen during pretraining (similar to the hold-out in \Cref{fig:exp_transfer_to_many:hold_out}). \textbf{Bold} denotes the best model on each dataset, \underline{underlined} the runner-up.
    }
    \label{tab:ucr_comparison_all_baselines}
    \begin{tabular}{lllllll}
    \toprule
    Target Dataset & XIT & TS-TCC & TF-C & TS2Vec & TNC & T-Loss \\
    \midrule
    ArrowHead & \underline{50.5 ±9.2} & 47.5 ±2.1 & 36.9 ±11.9 & \textbf{57.4 ±0.4} & 36.6 ±8.1 & 29.5 ±2.2 \\
    BeetleFly & \textbf{77.6 ±6.4} & 66.5 ±11.0 & 46.3 ±7.6 & \underline{68.0 ±5.9} & 67.2 ±14.5 & 61.7 ±6.9 \\
    BirdChicken & \underline{66.6 ±12.6} & 63.0 ±5.8 & 56.4 ±20.0 & \textbf{88.0 ±8.1} & 57.8 ±8.2 & 61.5 ±7.7 \\
    Car & 47.7 ±2.1 & \underline{60.2 ±10.6} & 25.4 ±6.0 & \textbf{73.0 ±2.2} & 29.9 ±14.1 & 23.5 ±6.0 \\
    Crop & \textbf{53.7 ±0.7} & \textbf{53.7 ±1.2} & \underline{47.1 ±0.8} & 40.4 ±2.5 & 19.6 ±9.8 & 14.1 ±1.1 \\
    Fish & \textbf{59.4 ±1.9} & \underline{55.0 ±7.9} & 31.8 ±6.9 & 44.6 ±7.3 & 12.5 ±6.3 & 08.5 ±4.6 \\
    FreezerRegularTrain & \textbf{77.2 ±0.3} & 76.2 ±0.3 & \underline{76.9 ±1.2} & \underline{76.9 ±0.5} & 73.1 ±8.0 & 58.3 ±3.0 \\
    GunPoint & 72.1 ±6.3 & \underline{78.9 ±2.1} & 58.7 ±17.0 & \textbf{94.4 ±2.8} & 62.4 ±22.4 & 64.6 ±12.0 \\
    GunPointMaleV.Female & \underline{92.7 ±2.2} & \textbf{95.7 ±3.7} & 76.2 ±4.8 & 86.7 ±3.4 & 75.5 ±8.0 & 74.3 ±3.8 \\
    Lightning7 & \underline{65.9 ±1.2} & 50.5 ±3.1 & 17.5 ±3.6 & \textbf{78.8 ±4.7} & 35.6 ±17.2 & 37.3 ±2.7 \\
    MedicalImages & \textbf{22.2 ±1.0} & 06.8 ±0.0 & \underline{12.2 ±3.0} & 06.8 ±0.0 & 09.7 ±4.3 & 08.2 ±2.5 \\
    MiddlePh.O.AgeGroup & \underline{31.8 ±4.7} & 20.2 ±7.3 & \textbf{39.6 ±1.8} & 21.1 ±2.3 & 23.9 ±13.1 & 25.4 ±14.9 \\
    MiddlePh.O.Correct & \textbf{36.3 ±0.0} & \textbf{36.3 ±0.0} & \textbf{36.3 ±0.0} & \textbf{36.3 ±0.0} & \textbf{36.3 ±0.0} & \textbf{36.3 ±0.0} \\
    MiddlePhalanxTW & \underline{19.9 ±3.7} & 14.3 ±2.3 & \textbf{24.3 ±2.0} & 15.6 ±0.7 & 14.3 ±7.4 & 14.6 ±8.1 \\
    OSULeaf & \underline{41.9 ±0.6} & 41.7 ±3.0 & 34.1 ±2.8 & \textbf{46.1 ±0.6} & 14.6 ±6.6 & 12.3 ±2.0 \\
    ProximalPh.O.Correct & 41.0 ±0.7 & 41.0 ±0.7 & \textbf{46.8 ±6.2} & 40.6 ±0.0 & \underline{41.4 ±1.8} & 40.6 ±0.0 \\
    ShapesAll & \textbf{70.9 ±0.2} & \underline{66.4 ±0.8} & 28.4 ±6.9 & 50.1 ±3.9 & 08.0 ±5.5 & 03.0 ±0.9 \\
    SonyAIBORobotSur.1 & 30.3 ±0.2 & 30.0 ±0.0 & \textbf{76.0 ±10.3} & 38.1 ±4.7 & \underline{47.8 ±13.1} & 45.7 ±14.0 \\
    SonyAIBORobotSur.2 & 51.6 ±11.8 & 50.9 ±11.6 & \textbf{86.2 ±1.1} & \underline{70.4 ±4.2} & 63.7 ±5.7 & 65.4 ±1.3 \\
    Symbols & 72.0 ±0.5 & \underline{74.7 ±1.3} & 36.7 ±3.0 & \textbf{90.7 ±1.9} & 45.1 ±19.5 & 53.6 ±16.7 \\
    Tiselac & \textbf{29.3 ±0.3} & \underline{26.3 ±0.7} & 19.4 ±0.5 & 18.5 ±1.2 & 15.3 ±2.7 & 12.3 ±0.2 \\
    ToeSegmentation2 & 69.6 ±7.4 & \underline{74.3 ±9.6} & 46.9 ±18.4 & \textbf{76.0 ±3.0} & 56.3 ±8.4 & 57.8 ±0.3 \\
    UWaveGestureLib.X & \textbf{77.1 ±0.3} & \underline{69.7 ±0.5} & 55.1 ±5.7 & 64.1 ±0.2 & 32.3 ±10.6 & 21.1 ±2.6 \\
    UWaveGestureLib.Z & \textbf{68.3 ±1.3} & \underline{61.3 ±1.9} & 51.4 ±3.2 & 55.5 ±1.1 & 27.1 ±11.8 & 22.9 ±3.1 \\
    WordSynonyms & \textbf{35.1 ±4.4} & \underline{17.3 ±0.6} & 06.6 ±1.1 & 03.3 ±0.5 & 04.0 ±1.3 & 03.1 ±1.7 \\
    \midrule
    Average Macro F1 $\uparrow$ & \textbf{54.4 ±3.2} & 51.1 ±3.5 & 42.9 ±5.8 & \underline{53.7 ±2.5} & 36.4 ±9.1 & 34.2 ±4.7 \\
    Rank $\downarrow$ & \textbf{2.10 ±1.2} & 3.06 ±1.6 & 3.56 ±1.8 & \underline{2.82 ±1.6} & 4.48 ±1.0 & 4.98 ±1.1 \\
    \bottomrule
    \end{tabular}
\end{table*}

\subsection{Ablation Studies: (Q3)}

\begin{table*}[t]
    \centering
    \caption{
        The ablation studies we performed to determine the key components driving the increase in performance observed earlier (Q3). We followed the same evaluation as in \Cref{tab:ucr_comparison_all_baselines}. Lower ranks are better, \textbf{bold} denotes best. The results demonstrate that each of the key components, XD-MixUp, TC loss, and SICC loss, contribute to the overall performance of XIT.
    }
    \label{tab:ablations}
    \begin{tabular}{llll}
    \toprule
    Pretraining Component & AUROC rank $\downarrow$ & Accuracy rank $\downarrow$ & Macro F1 rank $\downarrow$ \\
    \midrule
    XD-MixUp + SICC + TC (XIT) & \textbf{1.800 ±0.91} & \textbf{1.780 ±0.89} & \textbf{1.780 ±0.84} \\
    XD-MixUp + SICC & 3.560 ±0.87 & 3.700 ±0.56 & 3.580 ±0.79 \\
    XD-MixUp + TC & 2.060 ±0.94 & 1.920 ±0.84 & 1.980 ±0.91 \\
    TC & 2.580 ±0.91 & 2.600 ±0.78 & 2.660 ±0.81 \\
    \bottomrule
    \end{tabular}
\end{table*}

We conduct a series of ablation experiments to determine if the observed effects were due to the key components of XIT. We pretrained on the same three folds of 75 UCR datasets and finetuned on all 100 datasets. We ablated by systematically omitting our method's three key components XD-MixUp, TC loss, and SICC loss. Note that SICC is inherently linked to the presence of XD-Mixup, so SICC in isolation is not feasible. Similarly, only XD-MixUp without a loss is not a valid training procedure. The results are presented in \Cref{tab:ablations}. Our method XIT is superior to any ablated models when looking at the ranks for accuracy, AUROC, and Macro F1 scores. In particular, the high AUROC score means we have a high overall probability of correctly classifying classes, whereas the high Macro F1 score shows we can also correctly classify underrepresented classes.
Therefore, we conclude that all three building blocks of our combined pretraining procedure XIT are relevant for the performance increases observed in the previous section.

\subsection{Inspecting the Learned Representations: (Q4)}
\label{sec:experiments:inspect_latent}
\begin{figure}[t]
    \centering
    \includegraphics[width=\linewidth,trim={0 1.35cm 0 1.35cm},clip]{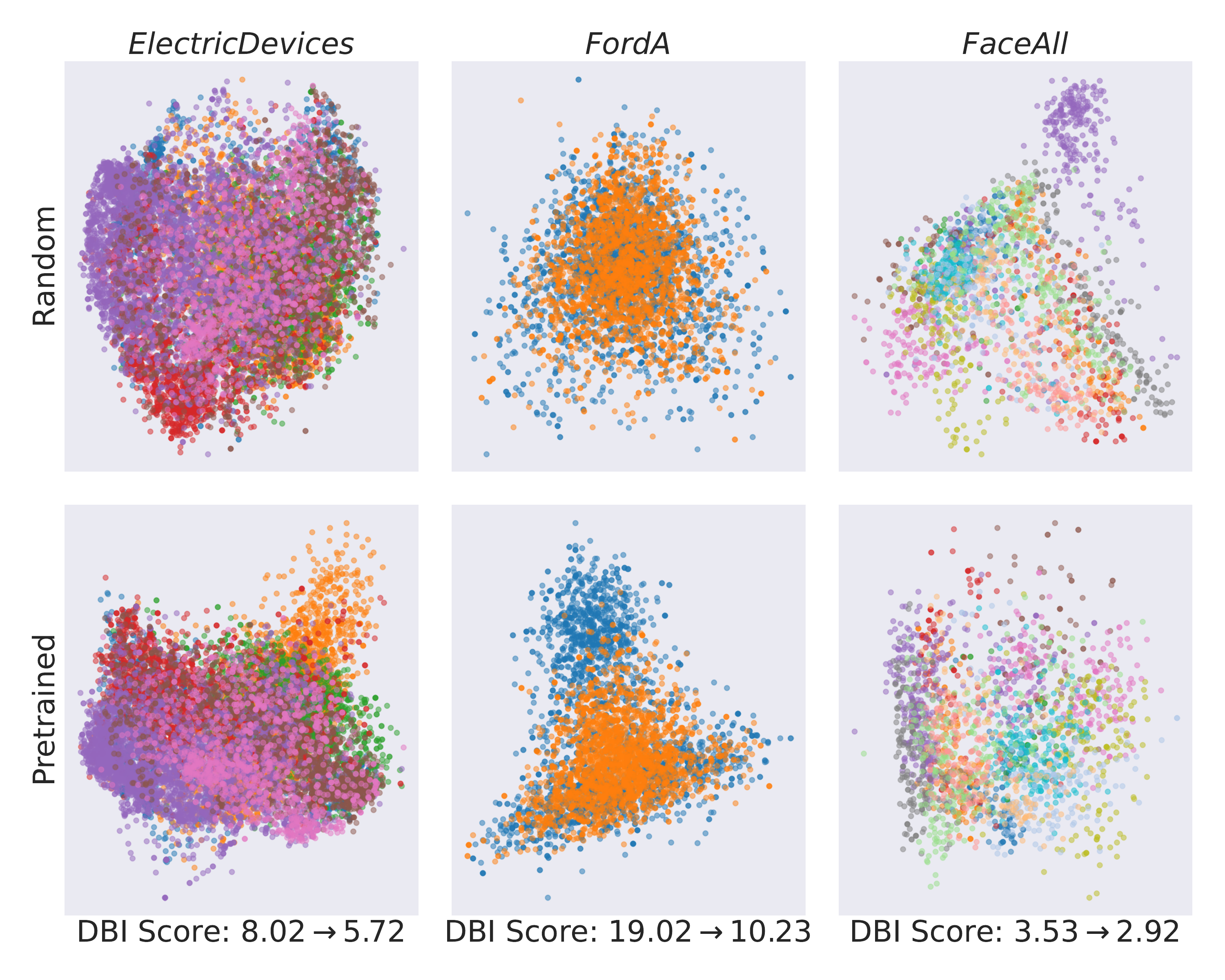}
    \caption{Low-dimensional embedding visualization from a single pretrained XIT model trained on 75 UCR datasets, evaluated on three hold-out datasets. Furthermore, we include the DBI to support our visual observation further. Lower DBI $\rightarrow$ more separated clusters.}
    \label{fig:pca_models}
\end{figure}

To gain insight into the effects of our proposed method, we examined the structure of multiple embedded time series datasets. We achieve a mean Davies-Bouldin Index~(DBI) \citep{daviesClusterSeparationMeasure1979} of 7.63 when embedded with a newly initialized encoder. After pretraining on 75 datasets, this score improved to 7.32.
This relatively small decrease suggests that our method does not need to alter the overall structure drastically but rather rearranges the intra-dataset structure. This effect can be seen qualitatively in \Cref{fig:pca_models}, which shows a projection of the latent representation learned by XIT of multiple unseen datasets. The figure was generated by performing a reduction to two dimensions via Principal Component Analysis on the output of the encoder $\mathcal{F}$. We observe a major structuring effect without any finetuning or the introduction of any labels. This demonstrates that in the majority of cases, the pretraining losses induce beneficial structure in the latent space, which is consistent with the results of \Cref{fig:exp_transfer_to_many}.

\section{Related Work}

Pretraining is a key element of current deep learning, allowing state-of-the-art outcomes in areas with scarce labels or data by utilizing a shared representation as the basis for adapting to the target domain. Although it has been extensively studied for domains such as natural language processing and computer vision, it still remains a challenge for the time series domain~\citep{maSurveyTimeSeriesPreTrained2023}. We can generally differentiate between supervised, unsupervised, and self-supervised pretraining. The core idea for the former is to utilize labels to steer the representation learning, while the latter two approaches work without any labels. Due to the wide availability of unlabeled time series datasets, we will focus on self-supervised methods in this paper.

\densePar{Self-Supervised Time Series Pretraining}
Several methods have been proposed to pretrain models on unlabeled datasets. The three main types of losses to optimize are reconstruction, pseudo-labels, and contrastive methods~\citep{maSurveyTimeSeriesPreTrained2023}.
The SimMTM framework~\citep{dongSimMTMSimplePreTraining2023} reconstructs time series from multiple masked variants and series-wise similarities within and across domains. Further models based on reconstructions are Ti-MAE~\citet{liTiMAESelfSupervisedMasked2023} and its extension TimeMAE~\citet{chengTimeMAESelfSupervisedRepresentations2023}.
Contrastive methods, on the other hand, need means to generate view pairs, e.g., as proposed in LEAVES~\citep{yuLEAVESLearningViews2022}, by \citet{tangExploringContrastiveLearning2021}, or in PAITS~\citep{beebe-wangPAITSPretrainingAugmentation2023}.
\citet{shiSelfSupervisedPretrainingTime2021} learn long-term dependencies using Dynamic Time Warping~(DTW) \citep{sakoeDynamicProgrammingAlgorithm1978}.
\citet{kiyassehCLOCSContrastiveLearning2021} specifically apply their contrastive learning method CLOCS to ECG signals.
In TS2Vec by \citet{yueTS2VecUniversalRepresentation2022}, instance-wise and temporal hierarchical contrasting is used to capture multiscale contextual information and dynamics.
TS-TCC~\citep{eldeleTimeSeriesRepresentationLearning2021} is a pretraining framework involving two views generated by weak and strong augmentations, which we also use in XIT. This is then fed into a temporal and contextual contrasting module using the NTXent loss~\citep{sohnImprovedDeepMetric2016} for learning robust and discriminative representations. Furthermore, the authors indicate that it is effective in transfer learning scenarios with few-labeled targets.
To use spectral information in contrasting, \citet{zhangSelfSupervisedContrastivePreTraining2022} developed \TFC{}, allowing the model to align the time and frequency domains with the respective views.
\citet{wickstromMixingContrastiveLearning2022} propose MixUp \newpage \citep{zhangMixupEmpiricalRisk2018} for the time domain, where two samples are combined by a sampled parameter $\lambda$, which is predicted as a pseudo-label. Furthermore, the two mixed views are aligned via the same contrastive NTXent loss used in TS2Vec, TS-TCC, \TFC{}, and our method XIT.
\citet{tonekaboniUnsupervisedRepresentationLearning2020} propose TNC, utilizing time series windows where ones with close-proximity share similar latent representations.
In addition, \cite{franceschiUnsupervisedScalableRepresentation2019a} propose T-Loss, which employs time-based negative sampling along with a triplet loss to learn an encoding.

\densePar{Multi-Dataset Pretraining}
While it is common to pretrain on a single source dataset, there is very little research in the area of multi-dataset pretraining, especially in the context of time series classification~\citep{maSurveyTimeSeriesPreTrained2023}. This arises from the fact that applying multiple datasets in a suboptimal setting may drastically decrease the performance~\citep{zhangSelfSupervisedContrastivePreTraining2022}, leading to a so-called \emph{negative transfer}.
Other works~\citep{gikundaHomogeneousTransferActive2021, tsengLightweightPretrainedTransformers2023,bruschMultiviewSelfsupervisedLearning2023} leverage multiple datasets in homogeneous settings where source and target distributions match.
Some works like OFA~\citep{zhouOneFitsAll2023}, Time-LLM~\citep{TimeLLMTimeSeries2023}, and LLMTime~\citep{gruverLargeLanguageModels2023} leverage existing large and diverse text datasets, however, doing so at the expense of using extremely resource-hungry large language models.
As far as we are aware,~\citet{kashiparekhConvTimeNetPretrainedDeep2019} and \citet{zhangSelfSupervisedContrastivePreTraining2022} are the only works that investigate proper multi-dataset pretraining. However, the former applies it in a supervised and very inflexible way, using one encoding head per source dataset, and the latter reports significant challenges when applying their method \TFC{} to multiple datasets at once, which they call \emph{many-to-one} setting. They note a clear drop in performance when increasing the number of datasets from one to two, three, and four.

\section{Conclusion \& Future Work}
Our research presents a paradigm shift in time series pretraining. Contrary to prevailing beliefs, our findings illustrate the possibility and effectiveness of multi-dataset pretraining for time series. By introducing XIT, consisting of XD-MixUp along with the SICC and TC losses, we have carved a promising path in self-supervised contrastive pretraining for time series. Our empirical evaluations showcased the efficacy of this method, especially in low-data regimes, against both supervised training and other self-supervised pretraining techniques. In essence, not only have we debunked the myth that multi-dataset pretraining is infeasible for time series, but we have also opened the door for further advancements in leveraging multiple datasets---beyond simultaneously using 75 datasets.

While our study has advanced time series pretraining, several promising directions beckon further exploration. The versatility of our approach needs evaluation on further tasks like forecasting and anomaly detection. We are eager to explore model reprogramming~\citep{yangVoice2SeriesReprogrammingAcoustic2021} to enhance adaptability and further decrease negative transfer. While we utilized MixUp augmentation, exploring specialized interpolations like DTW may yield further insights. Furthermore, we want to explore the potential of our SICC loss and integrated interpolation mechanism in other time series models and different modalities. Much like in NLP, future work might consider creating compound datasets with special attention to the types and distribution of the contained data.

\section*{Acknowledgments}
This work received funding by the EU project EXPLAIN, funded by the German Federal Ministry
of Education and Research (BMBF) (grant 01—S22030D).
It was further funded by the BMBF within the \enquote{The Future of Value Creation -- Research on Production, Services and Work} program (grant 02L19C150) managed by the Project Management Agency Karlsruhe~(PTKA).
Additionally, it was funded by the projects \enquote{The Adaptive Mind} and \enquote{The Third Wave of Artificial Intelligence -- 3AI} from the Hessian Ministry of Science and the Arts~(HMWK), the \enquote{ML2MT} project from the Volkswagen Stiftung, the ACATIS Investment KVG mbH project \enquote{Temporal Machine Learning for Long-Term Value Investing}, the ICT-48 Network of AI Research Excellence Center \enquote{TAILOR}~(EU Horizon 2020, GA No 952215), and the Collaboration Lab with Nexplore \enquote{AI in Construction}~(AICO).
The authors are responsible for the content of this publication.

\bibliography{references.bib}
\bibliographystyle{icml2024}

\onecolumn  %
\clearpage
\appendix
\section{Appendix}
\FloatBarrier

\subsection{Experimental Details}
\label{sec:app:experimental_setup}

This section gives more details for the experimental evaluation that should aid in reproducing our results. \Cref{tab:ucr_dataset_distribution} compares the datasets in the UCR repository we used, which are freely available from \url{https://timeseriesclassification.com}.

\begin{table*}
    \centering
    \caption{The distribution of datasets from the UCR repository used in our experiments.}
    \label{tab:ucr_dataset_distribution}
    \begin{tabular}{lrrrrrrrr}
    \toprule
    Domain & Datasets & \multicolumn{3}{c}{Sequence Length} & \multicolumn{3}{c}{Train Size} & Balanced \\
     &  & min & mean & max & min & median & max &  \\
    \midrule
    Audio & 2 & 270 & 337.5 & 405 & 60 & 132 & 204 & 50.0\% \\
    Device & 2 & 96 & 120.0 & 144 & 180 & 4553 & 8926 & 50.0\% \\
    Ecg & 4 & 82 & 113.5 & 140 & 23 & 61 & 500 & 50.0\% \\
    Eeg & 5 & 50 & 230.0 & 510 & 56 & 316 & 5890 & 40.0\% \\
    Har & 9 & 30 & 127.2 & 315 & 30 & 151 & 2238 & 55.6\% \\
    Image & 30 & 23 & 226.1 & 512 & 16 & 399 & 81714 & 20.0\% \\
    Meg & 1 & 200 & 200.0 & 200 & 727 & 727 & 727 & 0.0\% \\
    Motion & 14 & 8 & 229.8 & 343 & 36 & 332 & 7494 & 57.1\% \\
    Other & 2 & 36 & 118.5 & 201 & 18 & 1238 & 2459 & 0.0\% \\
    Sensor & 14 & 24 & 273.1 & 577 & 20 & 85 & 3636 & 42.9\% \\
    Simulated & 8 & 15 & 159.4 & 500 & 20 & 93 & 1000 & 75.0\% \\
    Sound & 1 & 217 & 217.0 & 217 & 3315 & 3315 & 3315 & 100.0\% \\
    Spectro & 7 & 234 & 382.0 & 570 & 28 & 57 & 613 & 28.6\% \\
    Traffic & 1 & 24 & 24.0 & 24 & 20 & 20 & 20 & 0.0\% \\
    \midrule
    Total & 100 & 8 & 221.1 & 577 & 16 & 180 & 81714 & 40.0\% \\
    \bottomrule
    \end{tabular}
\end{table*}

\paragraph{Implementation details} We used \emph{PyTorch}~\citep{paszkePyTorchImperativeStyle2019} version~2.0 as the base framework for all models. 
We performed almost all training in 16-bit mixed precision to save resources. For TS2Vec, we stayed very close to the reference implementation and therefore trained in 32-bit precision, used the existing hyperparameters for pretraining~(see \Cref{tab:hp}), and no early stopping in finetuning. Due to initially poor results with both TNC and T-Loss, we searched for optimal learning rates in $\{10^{-3}, 10^{-4}, 10^{-5}\}$~(results are in \Cref{tab:hp}). We used early stopping after four training epochs without improvements in the AUROC score for all other finetuning/supervised experiments. We ensured that we trained for at least 40 steps before stopping and up to a maximum of 2000 steps. The Adam optimizer with $\beta_1 = 0.9$ and $\beta_2 = 0.999$ was used to optimize all models with the hyperparameters given in \Cref{tab:hp} and gradient clipping to $1.0$. For the special case of the disproportionally large \emph{Tiselac} dataset, we increased the batch size to 256 for faster completion. For the data interpolation in \cref{eq:interpolation}, we set $\alpha = 0.2$ as determined by a hyperparameter search. In \cref{eq:total_loss} we set $\beta = 0.25$ according to our hyperparameter search, effectively giving three times more weight to $\mathcal{L}_\text{SICC}$ than to $\mathcal{L}_\text{TC}$. In $h(\bm c)$, $\BatchNorm1D$ normalizes the vectors per dimension by subtracting the mean and dividing by the empirical standard deviation. We used the default PyTorch configuration, maintaining a running average of the elements within the mini-batches. We set the temperature $\tau$ in \cref{eq:ell_sicc} to 0.2. We based our implementation of the baselines on the official repositories of:
\begin{itemize}
    \item TS-TCC~(\url{https://github.com/emadeldeen24/TS-TCC}),
    \item\TFC{}~(\url{https://github.com/mims-harvard/tfc-pretraining}),
    \item TNC~(\url{https://github.com/sanatonek/TNC_representation_learning}),
    \item T-Loss~(\href{https://github.com/White-Link/UnsupervisedScalableRepresentationLearningTimeSeries}{\texttt{https://github.com/White-Link/UnsupervisedScalableRepresentationLearning\\TimeSeries}}), and
    \item TS2Vec~(\url{https://github.com/yuezhihan/ts2vec}), respectively.
\end{itemize}

\begin{table*}[t]
    \centering
    \caption{This table shows the differing hyperparameters we chose by search using \emph{Optuna}~\citep{akibaOptunaNextgenerationHyperparameter2019}. In particular, we also considered larger batch sizes of up to 1,024 for the pretraining phase but did not find them to be beneficial, much like in the works of TS-TCC and \TFC{}.}
    \label{tab:hp}
    \begin{tabular}{@{}lllllll@{}}
    \toprule
               & \multicolumn{3}{c}{Pretraining}    & \multicolumn{3}{c}{Finetuning}      \\ \cmidrule(l){2-4} \cmidrule(l){5-7}
    Model      & Batch size & LR     & Weight decay & Batch size & LR      & Weight decay \\ \midrule
    XIT       & 64         & 0.0001 & 0.0003       & 64         & 0.00014 & 0.0016       \\
    TS-TCC     & 128        & 0.0003 & 0.0003       & 64         & 0.00014 & 0.0016       \\
    \TFC{}        & 64         & 0.0003 & 0.0005       & 64         & 0.0003  & 0.0003       \\
    TS2Vec        & 16         & 0.001 & 0.0005& 64         & 0.00014 & 0.0016       \\
 TNC& 64& 0.001& 0.0005& 64& 0.00014&0.0016\\
 T-Loss& 20& 0.001& 0.0005& 64& 0.00014&0.0016\\
    Supervised & --         & --     & --           & 64         & 0.00014 & 0.0016       \\ \bottomrule
    \end{tabular}
\end{table*}

\densePar{Encoder \& Summarization Model}
We used a simple and efficient encoder $\mathcal{F}$ with three residual convolution layers and configured it as in the work of~\citet{wangTimeSeriesClassification2017} and TS-TCC. For the model $\mathcal{S}$ used to obtain the context vectors $\bm c$, we used the exact same transformer model configuration as in TS-TCC: A token dimension of $64$ obtained from a linear projection with bias term, multi-head attention with four heads, and a total of four pre-norm transformer layers. The feedforward $\MLP$s each consist of two layers with hidden dimensions $64$, with a $\ReLU$ activation and subsequent dropout layer in between and a final dropout layer at the end. Both dropout probabilities were set to $10\%$. We employed a transformer model~\citep{vaswaniAttentionAllYou2017} $\mathcal{S}$ to calculate the context vectors $\bm{c}$, similar to TS-TCC. We observed in the supplementary implementation that~\citet{eldeleTimeSeriesRepresentationLearning2021} did not include positional encodings in their transformer tokens and thus used a set model instead of a sequence model. Nevertheless, we decided not to use it either since adding a sine-cosine positional encoding~\citep{vaswaniAttentionAllYou2017} did not noticeably affect the results.

\densePar{Finetuning}
To evaluate the utility of the learned representation for classification, we trained simple classifiers $\mathcal{C}$ on top of all encoder models $\mathcal{F}_{\bm\theta}$. This is embedded into the complete procedure as shown in \Cref{alg:combined_procedure}. We perform linear probing ~\citep{oordRepresentationLearningContrastive2019}, i.e., learning a single linear layer classifier $\mathcal{C}$ on top of the embeddings learned by $\mathcal{F}_{\bm\theta}$. The final output is transformed with a softmax, and the training criterion is cross-entropy. The hyperparameters for finetuning are given in \Cref{tab:hp}.

\paragraph{Details on \TFC{}} We changed the frequency transformation via the FFT to be orthonormal by scaling the result of length $T$ by $\sqrt{T}$. This preserves the signal's magnitude, allowing us to perform training and inference in 16-bit mixed precision without numerical issues. Furthermore, we use the complete pretraining datasets instead of only subsets for the N-to-one settings~\citep[Appendix K]{zhangSelfSupervisedContrastivePreTraining2022}. We use mostly the same hyperparameters as in Appendix E when pretraining \TFC{}. However, we deviate slightly to follow the linear probing evaluation. This means that in finetuning, we only train a single-layer classifier head instead of a deeper MLP. We only optimize the classifier and classifier loss instead of training the encoder as well or optimizing the pretraining and classifier losses jointly. We use the very same encoder config for all datasets for a more direct comparison, especially in the multi-dataset experiments.

\subsection{Reproducibility}
We acknowledge the significance of reproducibility in scientific research and have taken multiple steps to ensure the strength and replicability of our work. 

\begin{itemize}
    \item \textbf{Code:} We have used publicly available software and libraries to guarantee accessibility and have comprehensively described the architecture, software, versions, and hyperparameters in the~\Cref{sec:app:experimental_setup}. Our code is deterministic, incorporating seeds for all random number generators to guarantee the replicability of results. Furthermore, we will make our implementation available in a public repository at a later point in time for other researchers to access, review, and build upon.
\item \textbf{Datasets:} This study only utilizes publicly available datasets that have been correctly cited. Furthermore, the authors contribute to an open-source repository containing all the datasets used in this work, which will be made available upon acceptance.
\item \textbf{Architecture and Algorithm Details:} We have provided thorough descriptions and formulations of our architecture in the main text, supplemented by additional clarifications and implementation details in the~\Cref{sec:app:experimental_setup}, ensuring a clear understanding of our contributions and facilitating reproduction. This documentation is intended to provide researchers with all the necessary information to replicate our experiments accurately.
\end{itemize}

\end{document}